\documentclass[sigconf]{acmart}

\usepackage{microtype}
\usepackage{graphicx}
\usepackage{subfigure}
\usepackage{booktabs} 
\usepackage[ruled,linesnumbered]{algorithm2e}
\usepackage{amsfonts}
\usepackage{hhline}

\usepackage{hyperref}
\usepackage{threeparttable}
\usepackage{dblfloatfix}

\usepackage{bm}
\usepackage{amsmath}
\usepackage[T1]{fontenc}
\definecolor{deepblue}{RGB}{31,119,180}

\AtBeginDocument{%
  \providecommand\BibTeX{{%
    \normalfont B\kern-0.5em{\scshape i\kern-0.25em b}\kern-0.8em\TeX}}}

\begin{document}

\title{Less Is More: Fast Multivariate Time Series Forecasting with Light Sampling-oriented MLP Structures}


\author{Tianping Zhang}
\affiliation{%
  \institution{Tsinghua University}
  \city{Beijing}
  \country{China}
}
\email{ztp18@mails.tsinghua.edu.cn}

\author{Yizhuo Zhang}
\affiliation{%
  \institution{Tsinghua University}
  \city{Beijing}
  \country{China}
}
\email{zyz22@mails.tsinghua.edu.cn}

\author{Wei Cao}
\affiliation{%
  \institution{Microsoft Research of Asia}
  \city{Beijing}
  \country{China}
}
\email{weicao@microsoft.com}

\author{Jiang Bian}
\affiliation{%
  \institution{Microsoft Research of Asia}
  \city{Beijing}
  \country{China}
}
\email{jiang.bian@microsoft.com}

\author{Xiaohan Yi}
\affiliation{%
  \institution{Microsoft Research of Asia}
  \city{Beijing}
  \country{China}
}
\email{xiaoyi@microsoft.com}

\author{Shun Zheng}
\affiliation{%
  \institution{Microsoft Research of Asia}
  \city{Beijing}
  \country{China}
}
\email{shun.zheng@microsoft.com}

\author{Jian Li}
\affiliation{%
  \institution{Tsinghua University}
  \city{Beijing}
  \country{China}
}
\email{lijian83@mail.tsinghua.edu.cn}


\begin{abstract}
Multivariate time series forecasting has seen widely ranging applications in various domains, including finance, traffic, energy, and healthcare. To capture the sophisticated temporal patterns, plenty of research studies designed complex neural network architectures based on many variants of RNNs, GNNs, and Transformers. However, complex models are often computationally expensive and thus face a severe challenge in training and inference efficiency when applied to large-scale real-world datasets. In this paper, we introduce LightTS, a light deep learning architecture merely based on simple MLP-based structures. The key idea of LightTS is to apply an MLP-based structure on top of two delicate down-sampling strategies, including {\em interval sampling} and {\em continuous sampling}, inspired by a crucial fact that down-sampling time series often preserves the majority of its information. We conduct extensive experiments on eight widely used benchmark datasets. Compared with the existing state-of-the-art methods, LightTS demonstrates better performance on five of them and comparable performance on the rest. Moreover, LightTS is highly efficient. It uses less than $5\%$ FLOPS compared with previous SOTA methods on the largest benchmark dataset. In addition, LightTS is robust and has a much smaller variance in forecasting accuracy than previous SOTA methods in long sequence forecasting tasks. Our codes and datasets are available in the anonymous link.\footnote{\url{https://tinyurl.com/5993cmus}}


\end{abstract}




\maketitle

\section{Introduction}
Multivariate time series, as one of fundamental real-word data types, consists of more than one time-dependent variable, and, more importantly, each variable depends not only on its past values but also has some dependency on other variables. 
Multivariate time series forecasting has become a critical application task in various domains, including economics, traffic, energy, and healthcare. The vital modeling challenge lies in capturing 1) sophisticated temporal patterns (both short-term local patterns and long-term global patterns) of each variable as well as 2) complex inter-dependency among different variables. Due to the ability of deep learning to model complex patterns, especially its successful applications in computer vision and natural language processing tasks, there have been significantly growing research interests to apply deep neural networks into multivariate time series forecasting \cite{N-Beats} rather than traditional methods (such as ARIMA \cite{box2015time}, Holt-Winters \cite{Holt-winters}, etc). 

Particularly, with increasing computing power and the development of neural network architectures, many recent studies turned their eyes upon the realms of RNNs, GNNs, and Transformers. For instance, LSTNet \cite{LSTNet} and TPR-LSTM \cite{TPR-LSTM} used the hybrids of Convolutional Neural Networks (CNNs), Recurrent Neural Networks (RNNs), and the Attention Mechanism \cite{attention} to capture the long/short-term dependencies. Informer \cite{Informer} and Autoformer \cite{Autoformer} further explored the potential of Transformer in capturing very long-range dependency of time series. 
MTGNN \cite{MTGNN}, StemGNN \cite{StemGNN}, and IGMTF \cite{IGMTF} used Graph Neural Networks (GNNs) to explicitly model the dependencies among variables. While all these proposed complex deep learning models have demonstrated promising performance under specific scenarios, the introduced sophisticated neural network architectures usually indicate computationally expensive training and inference process, especially when facing the time series with long input length and many multiple correlated variables. 
Moreover, complex neural network architectures are always data-hungry due to the large number of parameters~\cite{kitaev2020reformer, Informer}, which may cause the trained models not robust when the amount of available training data is limited. 
Motivated by the challenges above, we naturally raise a question, {\em is it necessary to apply complex and computationally expensive models to achieve state-of-the-art performance in multivariate time series forecasting?}

In this paper, we explore the possibility of using simple and lightweight neural network architectures, i.e., merely using simple multi-layer perceptron (MLP) structure, for accurate multivariate time series forecasting. Specifically, relying on a critical observation that down-sampling time series often preserves the majority of its information \cite{SCINet} due to the trending, seasonal and irregular characteristics of time series, we thus propose LightTS, a light deep learning architecture constructed exclusively based on simple MLP-based structures. The key idea of LightTS is to apply an MLP-based structure on top of two delicate down-sampling strategies, including {\em interval sampling} and {\em continuous sampling}, inspired by crucial characteristics of multivariate time series. To be more concrete, continuous sampling focuses on capturing the short-term local patterns, while interval sampling focuses on capturing the long-term dependencies. On top of the sampling strategies, we propose an MLP-based architecture that can exchange information among different down-sampled sub-sequences and time steps. In such a way, the model can adaptively select useful information for forecasting from both local and global patterns. Furthermore, since each time the model only needs to deal with a fraction of the input sequence after down-sampling, our model is highly efficient in handling time series with very long input length.

The contributions of this paper are summarized as follows:

\begin{itemize}
    \item We propose LightTS, a simple architecture that is highly efficient and accurate in multivariate time series forecasting tasks. To the best of our knowledge, this is the first work that demonstrates the great potential of the MLP-based structures in multivariate time series forecasting.
    \item We propose continuous and interval sampling according to the special property of time series. The sampling methods help capture long/short-term patterns effectively and enable greater efficiency for long input sequences.
    \item Experimental results show that LightTS is competent in both short sequence and long sequence forecasting tasks. LightTS outperforms the state-of-the-art methods on 5 out of 8 widely-used benchmark datasets. In addition, LightTS is highly efficient. LightTS uses less than 5\% FLOPS on the largest benchmark dataset compared with previous SOTA methods. Moreover, LightTS is robust and has a much smaller variance in forecasting accuracy than previous SOTA methods in long sequence forecasting tasks.
\end{itemize}

We hope that our results can inspire further research beyond the realms of RNNs, GNNs, and Transformers in time series forecasting. We organize the rest of the paper as follows. We start with a survey of related work in Section \ref{sec: related work}. We formally define our problem in \ref{sec: problem definition}, and we present our model LightTS in Section \ref{sec: model}. 
We present experimental results in Section \ref{sec: Experiments}. We provide discussion in Section \ref{sec: discussion}. We conclude our paper in Section \ref{sec: conclusion}.

\section{Related Work}
\label{sec: related work}
Time series forecasting methods can generally be classified into statistical and deep-learning-based methods.
\subsection{Statistical Methods}
Statistical methods for time series forecasting have been studied for a long time. Traditional methods include auto-regression (AR), moving average (MA), and auto-regressive moving average (ARMA). The auto-regressive integrated moving average model (ARIMA) \cite{box2015time} extends the ARMA model by incorporating the notion of integration. The vector autoregressive model (VAR) \cite{box2015time} is an extension of the AR model and captures the linear dependencies among different time series. Gaussian Process (GP) \cite{GP} is a Bayesian approach that models the distribution of multivariate time series over continuous functions. Statistical methods are popular for their simplicity and interpretability. However, those approaches require strong assumptions (such as the stationary assumption) and do not scale well to large multivariate time series data.

\subsection{Deep-learning-based Methods}
Recently, deep-learning-based methods have become increasingly popular in time series forecasting. LSTNet \cite{LSTNet} and TPR-LSTM \cite{TPR-LSTM} both employ the convolutional neural networks, the recurrent neural networks, and the attention mechanism \cite{attention} to model short-term local dependencies among different time series and long-term temporal dependencies. MTGNN \cite{MTGNN}, StemGNN \cite{StemGNN}, and IGMTF \cite{IGMTF} leverage the graph neural networks to explicitly model the correlations among different time series. SCINet \cite{SCINet} downsamples the time series and uses convolutional filters to extract features and interact information. VARMLP \cite{VARMLP} is a hybrid model that combines the AR model and an MLP with a single hidden layer, which is substantially different from LightTS we propose.

Long sequence time series forecasting (LSTF) has received increasing attention in the forecasting community. LSTF is important for various real-world applications (e.g., energy consumption planning). RNN-based models such as LSTNet \cite{LSTNet} have limited prediction capacity in LSTF \cite{Informer}. Informer \cite{Informer} improves the vanilla Transformer \cite{attention} in time complexity and memory usage for LSTF tasks. Autoformer \cite{Autoformer} uses decomposition and an auto-correlation mechanism to achieve better results. We show through experiments that LightTS is more efficient and effective for long sequence time series forecasting. 

\begin{figure*}[ht]
    \centering
    \includegraphics[width=0.95\textwidth]{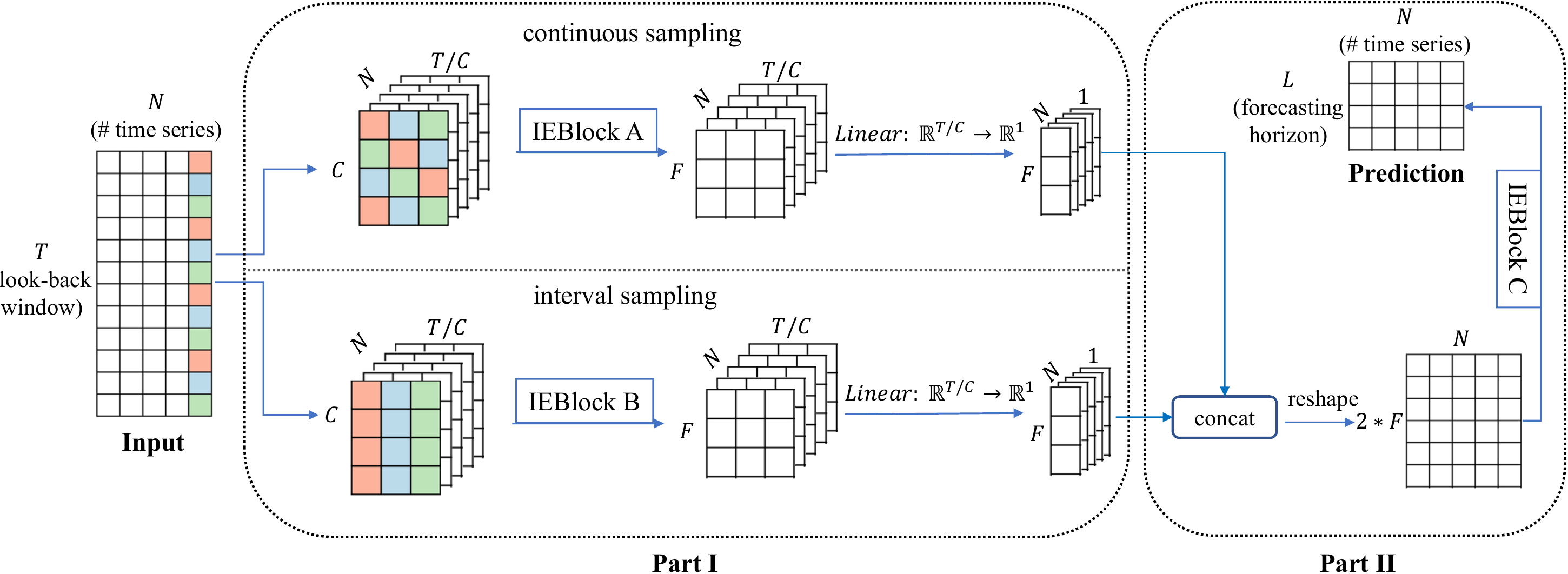}
    \caption{The overview of LightTS. In Part I, the model captures the short/long-term dependencies and extract features of each time series. In Part II, the model learns the interdependencies among different time series and make predictions.}
    \label{fig:LightTS}
\end{figure*}

Proposing a pure MLP-based model for multivariate time series forecasting is partly motivated by N-BEATS \cite{N-Beats}. N-BEATS is the first pure deep-learning-based method for univariate time series forecasting that achieves SOTA in the M4 competition\footnote{The M competitions \cite{M4, M5} are famous forecasting competitions that focus on univariate time series forecasting. N-BEATS is proposed after the competition.} \cite{N-Beats, M4}. In the ablation study of N-BEATS, where the backward residual connection is disabled, the architecture becomes a pure MLP-based structure. The architecture shows comparable results to N-BEATS and the winning solution of the M4 competition\footnote{The MLP-based architecture has an OWA of 0.822 in the M4 dataset, where OWA of the winning solution in the M4 competition is 0.821 and the second best is 0.838 \cite{M4}. OWA (overall weighted average of sMAPE and MASE) is the major accuracy measure of the M4 competition \cite{M4}.}. This finding supports the fact that, with careful design, MLP-based structures are powerful in capturing the historical patterns of time series.

Recently, several MLP-based architectures have been proposed for computer vision, such as MLP-Mixer \cite{MLP-Mixer}, gMLP \cite{gMLP}, and ResMLP \cite{ResMLP}. Such architectures leverage information exchange over channels and spatial tokens. Compared with those architectures, our model is different since we consider the information exchange both in the original input and down-sampled sub-sequences.

\section{Our Model: LightTS}
\subsection{Problem Definition}
\label{sec: problem definition}
We first formulate the problem of multivariate time series forecasting. At timestamp {\em t}, given a look-back window of fixed length {\em T}, we have a series of observations $X_t=\{x_{t-T+1}, ..., x_{t-1}, x_t|x_i\in \mathbb{R}^N\}$ where $N$ is the number of time series (i.e., variables in a multivariate sample). Given a forecasting horizon {\em L}, our goal is to predict either the values on multiple timestamps $\{x_{t+1},...,x_{t+L}\}$, which is called multi-step forecasting; or the value of $x_{t+L}$, which is called single-step forecasting. Long sequence time series forecasting usually has $L$ much larger than one hundred \cite{Informer}.

\subsection{Architecture Overview}
\label{sec: model}
We present the overall architecture of LightTS in  Figure \ref{fig:LightTS}. 
Recall that there are two major challenges in multivariate time series forecasting: (1) capture the short-term local patterns and long-term global patterns; (2) capture the interdependencies among different time series variables. 
LightTS also consists of two parts that correspond to the two challenges. In the first part, we treat different time series (i.e., input variables) independently without considering their interdependencies. This part aims to capture the short/long-term dependencies and extract the corresponding features of each time series (the first challenge). In the second part, we concatenate all the time series and learn the correlations among different input variables (the second challenge).

The key components of these two parts are two sampling methods called continuous sampling and interval sampling, and three {\em Information Exchange Block (IEBlock)}, where we describe the details in the following subsections.

\subsection{Continuous Sampling and Interval Sampling}
We first present the sampling strategies used in LightTS. Compared with the other sequential data such as natural language and audio data, time series is a special sequence in that down-sampling the time series often preserves the majority of its information \cite{SCINet}. Nevertheless, the na\"ive down-sampling method (such as the uniform sampling) could lead to information loss. Motivated by this, we design continuous and interval sampling, which helps the model capture both the local and global temporal patterns without eliminating any tokens.

\begin{figure*}[thp]
    \centering
    \includegraphics[width=17.5cm]{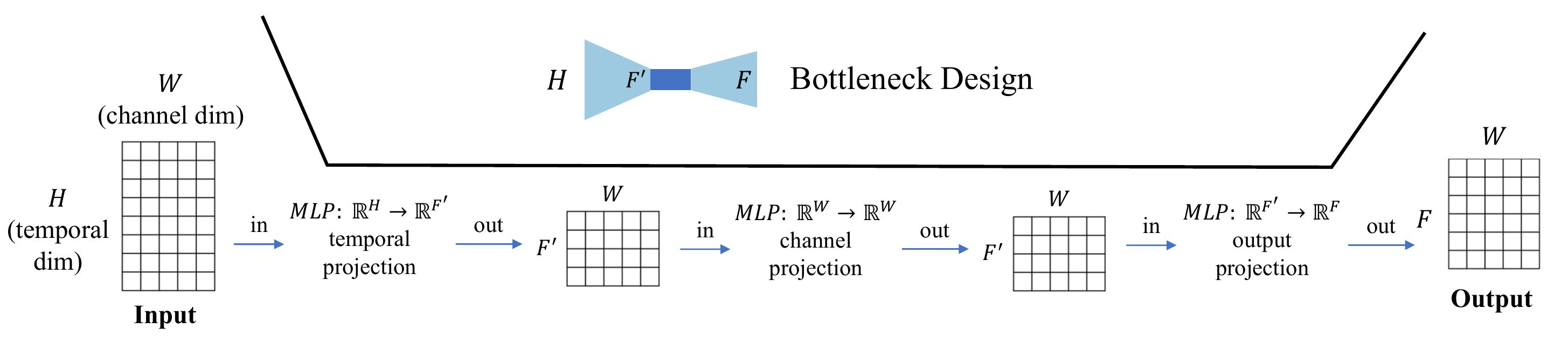}
    \caption{The overview of IEBlock and the bottleneck design.}
    \label{fig: IEBlock}
\end{figure*}

Our sampling methods transform each time series of length $T$ to several non-overlapping sub-sequence with length $C$. For continuous sampling, each time we consecutively select $C$ tokens as one sub-sequence. Thus, for a input sequence $X_t \in \mathbb{R}^T$, we down-sample it to $\frac{T}{C}$ sub-sequences and obtain a $2D$ matrix $X^{con}_t \in \mathrm{R}^{C \times \frac{T}{C}}$, where the $j$-th column is 
\begin{equation}
{X^{con}_t}_{\bm \cdot j} = \{x_{t - T + (j - 1) \cdot C + 1},  x_{t - T + (j - 1) \cdot C + 2}, \ldots, x_{t - T + j \cdot C}\}
\end{equation}
For interval sampling, each time we sample $C$ tokens with a fixed time interval. Similar to continuous sampling, the $j$-th column of the down-sample matrix for interval sampling $X^{int}_t$ is
\begin{equation}
    {X^{int}_t}_{\bm \cdot j} = \{x_{t - T + j}, x_{t - T + j + \lfloor \frac{T}{C} \rfloor}, x_{t - T + j + 2 \cdot \lfloor \frac{T}{C} \rfloor}, \ldots, x_{t-T + j+(C-1)\cdot \lfloor \frac{T}{C} \rfloor}\}
\end{equation}
The proposed sampling methods help the models focus on specific temporal patterns. For example, for continuous sampling, we sub-sample the time series into continuous pieces without long-range information, thus the model would pay more attention to the short-term local patterns. For interval sampling, we put aside local details, and the model would focus on long-term global patterns. Such design makes the model more effective and efficient to train since we only need to handle a fraction of the input sequence, especially when the input sequence is very long. We emphasize that unlike the na\"ive down-sampling methods, we do not eliminate any tokens here. Instead, we keep all the original tokens and transform them into several non-overlapping sub-sequences. The following section presents an MLP-based architecture to learn the useful features from the down-sampled sub-sequences.

\subsection{Information Exchange Block}
Information Exchange Block (IEBlock) is the basic building block we design for LightTS. In short, an IEBlock takes a $2D$ matrix of shape $H \times W$, where $H$ is the temporal dimension and $W$ is the channel dimension. The goal of an IEBlock is to leverage the information exchange along different dimensions and outputs another feature map of shape $F \times W$ ($F$ is the hyperparameter that corresponds to the output feature dimension). The obtained matrix can be regarded as the extracted features of the input matrix. IEblocks are the key components of LightTS and are used in both the down-sampling part and the prediction part.

We present the architecture of IEBlock in Figure \ref{fig: IEBlock}. We use $Z=(z_{ij})_{H\times W}$ to denote the input 2D matrix, $\bm{z_{\cdot i}}=(z_{1i}, z_{2i},...,z_{Hi})^T$ to denote the $i$-th column and $\bm{z_{j\cdot}}=(z_{j1},z_{j2},...,z_{jW})$ to denote the $j$-th row. 
We first apply an MLP of $\mathbb{R}^{H} \rightarrow \mathbb{R}^{F^{'}}$ on each column, where $F' \ll F$.
We refer to such operation as the temporal projection:
$$
\bm{z_{\cdot i}^{t}}=\mathrm{MLP}(\bm{z_{\cdot i}}), \forall i=1,2,...,W
$$
The temporal projection extracts features along the temporal dimension. We use weight sharing for all columns in the temporal projection for efficiency. Next, we apply an MLP of $\mathbb{R}^W \rightarrow \mathbb{R}^W$ on each row, which we refer to as the channel projection:
$$
\bm{z_{j\cdot}^{c}}=\mathrm{MLP}(\bm{z_{j\cdot}^{t}}), \forall j=1,2,...,F^{'}.
$$
The channel projection exchanges information among different channels but keeps the input shape unchanged. We also use weight sharing for all rows in the channel projection. Finally, we apply another MLP of $\mathbb{R}^{F'} \rightarrow \mathbb{R}^F$ on each column to map the feature dimension from $F'$ to $F$,
which we refer to such operation as the output projection:
$$
\bm{z_{\cdot i}^{o}}=\mathrm{MLP}(\bm{z_{\cdot i}}^{c}), \forall i=1,2,...,W
$$
We call such an architecture the ``bottleneck architecture", where the middle layer feature dimension is far less than the output layer feature dimension.

Note that we use IEBlocks in both the down-sampling part and prediction part. In different parts, $H$ and $W$ could have different meanings.
For example, in the sampling part, $H$ corresponds to the length of sub-sequence $C$, and $W$ corresponds to the number of sub-sequences $\frac{T}{C}$. In the prediction part, $H$ corresponds to the feature dimension extracted in the first part, and $W$ corresponds to the number of time series variables $N$. 

In an IEBlock, we repeatedly apply the channel projection on each time step. When the input time series is long, such an operation leads to expensive computational costs. Therefore, we propose a bottleneck design in IEBlock. First, we use a temporal projection to map the number of rows from $H$ to $F^{'}$. Then, the channel projection is applied $F^{'}$ times to communicate information. Finally, we use an output projection to map the features $F^{'}$ to the desired output length $F$. We call this bottleneck design since the hyperparameter $F^{'}$ is much smaller than $H$ and $F$.

\subsection{Training Procedure}
Now, we can describe the training procedure of LightTS. In the first part, we transform each time series with length $T$ to a $2D$ matrix with shape $C \times \frac{T}{C}$. We then use two IEBlocks (IEBlock-A and IEBlock-B in the Figure \ref{fig:LightTS}) to extract the corresponding temporal features and obtain the feature matrices with shape $\mathbb{R}^{F \times \frac{T}{C}}$. Each feature matrix is then down-projected to $\mathbb{R}^F$ with a simple linear mapping $\mathbb{R}^{\frac{T}{C}} \rightarrow \mathbb{R}$. Hence, we map each time series to a feature vector of dimension $F$. This part focuses on capturing the long/short-term temporal patterns without considering the correlations of different time series variables.

In the second part, we concatenate all the obtained features in the first part. We concatenate the features from continuous and interval sampling on the temporal dimension and all the time series variables on the channel dimension. We thus obtain an input matrix of shape $\mathbb{R}^{2F \times N}$. Finally, we introduce another IEBlock-C of $\mathbb{R}^{2F \times N} \rightarrow \mathbb{R}^{L \times N}$. IEBlock-C aims to combine the short-term local patterns and long-term global patterns from the temporal dimension and the correlations among different input variables from the channel dimension. The output of IEBlock-C is our final prediction.

\section{Experiments}
\label{sec: Experiments}
We validate LightTS on eight public benchmark datasets, from short sequence forecasting to long sequence forecasting. Following the previous studies \cite{LSTNet, Informer}, we use the single-step setting for short sequence forecasting and the multi-step setting for long sequence forecasting. We demonstrate the advantages of LightTS over previous methods in terms of accuracy, efficiency, and robustness, respectively. 

\begin{table}[th]
  \centering
  \small
  \caption{Dataset statistics}
    \begin{tabular}{|c|cccc|}
    \hline
    Datasets & Variants & Timesteps & Granularity & Start time \\
    \hline
    ETTh1 & 7     & 17420 & 1 hour & 2016/7/1 \\
    ETTh2 & 7     & 17420 & 1 hour & 2016/7/1 \\
    ETTm1 & 7     & 69680 & 15 minutes & 2016/7/1 \\
    Weather & 12    & 35064 & 1 hour & 2010/1/1 \\
    Solar-Energy & 137   & 17544 & 10 minutes & 2006/1/1 \\
    Traffic & 862   & 52560 & 1 hour & 2015/1/1 \\
    Electricity & 321   & 26304 & 1 hour  & 2012/1/1 \\
    Exchange-Rate & 8     & 7588  & 1 day & 1990/1/1 \\
    \hline
    \end{tabular}%
  \label{tab:dataset}%
\end{table}%

\subsection{Experimental Settings}
\subsubsection{Datasets}
\label{sec: datasets}
In Table \ref{tab:dataset}, we summarize the statistics of eight benchmark datasets. Following the experimental setting in the previous studies \cite{Informer, SCINet}, for short sequence forecasting, we use the  {\em Solar-Energy}, {\em Traffic}, {\em Electricity} and {\em Exchange-Rate} datasets \cite{LSTNet, MTGNN}; for long sequence forecasting, we use the {\em Electricity Transformer Temperature} (ETTh1, ETTh2, ETTm1), {\em Electricity} and {\em Weather} datasets \cite{Informer}. We provide more details about the datasets in Appendix \ref{app: datasets}.

\subsubsection{Evaluation Metrics}
\label{sec: evaluation metrics}
For a fair comparison, we use the same evaluation metrics as previous studies \cite{LSTNet, Informer}. We use the Mean Squared Error (MSE) and Mean Absolute Error (MAE) for long sequence forecasting. We use the Root Relative Squared Error (RSE) and Empirical Correlation Coefficient (CORR) for short sequence forecasting. Readers can find more detailed information in Appendix \ref{app:evaluation}.

\subsection{Baseline Methods for Comparisons}
We summarize the baseline methods in the following:
\subsubsection{Short sequence forecasting}
\begin{itemize}
    \item AR: An auto-regressive model.
    \item VARMLP \cite{VARMLP}. A hybrid model that combines the AR model and an MLP.
    \item GP \cite{GPTS}. A Gaussian Process model.
    \item RNN-GRU \cite{MTGNN}. A recurrent neural network with GRU hidden units.
    \item LSTNet \cite{LSTNet}. A hybrid model that combines convolutional neural networks and recurrent neural networks.
    \item TPR-LSTM \cite{TPR-LSTM}. A hybrid model that combines recurrent neural networks and the attention mechanism.
    \item TCN \cite{TCN}. A typical temporal convolutional network.
    \item SCINet \cite{SCINet}. A forecasting model with sample convolution and interaction.
    \item MTGNN \cite{MTGNN}. A GNN-based method that explicitly models the correlations among different time series.
\end{itemize}

\subsubsection{Long sequence forecasting}
In addition to LSTNet and SCINet, we compare with the following baseline methods for long sequence forecasting:
\begin{itemize}
    \item LogTrans \cite{LogTrans}. A variant of Transformers with LogSparse self-attention mechanism.
    \item Reformer \cite{kitaev2020reformer}. An efficient variant of Transformers with locality-sensitive hashing.
    \item LSTMa \cite{LSTMa}. A variant of recurrent neural networks with dynamic length of encoding vectors.
    \item Informer \cite{Informer}. A variant of Transformers with ProbSparse self-attention mechanism.
    \item Autoformer \cite{Autoformer}. A variant of Transformers with a decomposition forecasting architecture.
\end{itemize}

\begin{table*}[!t]\scriptsize
  \centering
  \renewcommand\arraystretch{1.6}
  \caption{Baseline comparisons under multi-step setting for long sequence time series forecasting tasks.}
  
  \begin{threeparttable}
    \begin{tabular}{|c|c|ccc|ccc|ccc|ccc|ccc|}
    \hline
    \multicolumn{1}{|c}{Methods} & \multicolumn{1}{|c|}{Metrics} & \multicolumn{3}{c|}{ETTh1} & \multicolumn{3}{c|}{ETTh2} & \multicolumn{3}{c|}{ETTm1} & \multicolumn{3}{c|}{Weather} & \multicolumn{3}{c|}{Electricity} \\
\cline{3-17}          &       & \multicolumn{3}{c|}{horizon} & \multicolumn{3}{c|}{horizon} & \multicolumn{3}{c|}{horizon} & \multicolumn{3}{c|}{horizon} & \multicolumn{3}{c|}{horizon} \\
\cline{3-17}          &       & 168   & 336   & 720   & 168   & 336   & 720   & 96    & 288   & 672   & 168   & 336   & 720   & 336   & 720   & 960 \\
    \hline
    \multicolumn{1}{|c|}{LogTrans} & MSE   & 0.888  & 0.942  & 1.109  & 3.944  & 3.711  & 2.817  & 0.674  & 1.728  & 1.865  & 0.649  & 0.666  & 0.741  & 0.305  & 0.311  & 0.333  \\
          & MAE   & 0.766  & 0.766  & 0.843  & 1.573  & 1.587  & 1.356  & 0.674  & 1.656  & 1.721  & 0.573  & 0.584  & 0.611  & 0.395  & 0.397  & 0.413  \\
    \hline
    \multicolumn{1}{|c|}{Reformer} & MSE   & 1.686  & 1.919  & 2.177  & 4.484  & 3.798  & 5.111  & 1.267  & 1.632  & 1.943  & 1.228  & 1.770  & 2.548  & 1.507  & 1.883  & 1.973  \\
          & MAE   & 0.996  & 1.090  & 1.218  & 1.650  & 1.508  & 1.793  & 0.795  & 0.886  & 1.006  & 0.763  & 0.997  & 1.407  & 0.978  & 1.002  & 1.185  \\
    \hline
    \multicolumn{1}{|c|}{LSTMa} & MSE   & 1.058  & 1.152  & 1.682  & 3.987  & 3.276  & 3.711  & 1.195  & 1.598  & 2.530  & 0.948  & 1.497  & 1.314  & 0.778  & 1.528  & 1.343  \\
          & MAE   & 0.725  & 0.794  & 1.018  & 1.560  & 1.375  & 1.520  & 0.785  & 0.952  & 1.259  & 0.713  & 0.889  & 0.875  & 0.629  & 0.945  & 0.886  \\
    \hline
    \multicolumn{1}{|c|}{LSTNet} & MSE   & 1.865  & 2.477  & 1.925  & 1.442  & 1.372  & 2.403  & 2.654  & 1.009  & 1.681  & 0.676  & 0.714  & 0.773  & 0.357  & 0.442  & 0.473  \\
          & MAE   & 1.092  & 1.193  & 1.084  & 2.389  & 2.429  & 3.403  & 1.378  & 1.902  & 2.701  & 0.585  & 0.607  & 0.643  & 0.391  & 0.433  & 0.443  \\
    \hline
    \multicolumn{1}{|c|}{Informer} & MSE   & 0.878  & 0.884  & 0.941  & 1.512  & 1.665  & 2.340  & 0.642  & 1.219  & 1.651  & 0.592  & 0.623  & 0.685  & 0.311  & 0.308  & 0.328  \\
          & MAE   & 0.722  & 0.753  & 0.768  & 0.996  & 1.035  & 1.209  & 0.626  & 0.871  & 1.002  & 0.531  & 0.546  & 0.575  & 0.385  & 0.385  & 0.406  \\
    \hline
    \multicolumn{1}{|c|}{Autoformer*} & MSE   & 0.634  & 0.724  & 0.898  & 1.101  & 1.386  & 2.445  & 0.539  & 0.575  & \textcolor{deepblue}{\em 0.599 } & \textbf{0.359 } & \textbf{0.492 } & \textbf{0.527 } & 0.257  & 0.259  & 0.291  \\
          & MAE   & 0.590  & 0.651  & 0.743  & 0.803  & 0.892  & 1.226  & 0.504  & 0.527  & \textcolor{deepblue}{\em 0.542 } & \textbf{0.413 } & \textbf{0.491 } & \textbf{0.503 } &  0.357  &  0.361  & 0.381  \\
    \hline
    \multicolumn{1}{|c|}{SCINet*}   &  MSE   & \textcolor{deepblue}{\em 0.450 } & \textcolor{deepblue}{\em 0.528 } & \textcolor{deepblue}{\em 0.597 } & \textcolor{deepblue}{\em 0.554 } & \textcolor{deepblue}{\em 0.657 } & \textcolor{deepblue}{\em 1.118 } & \textcolor{deepblue}{\em 0.197 } & \textcolor{deepblue}{\em 0.350 } & 1.214  & 0.515  & 0.540  & 0.577  & \textcolor{deepblue}{\em 0.198 } & \textcolor{deepblue}{\em 0.234 } & \textcolor{deepblue}{\em 0.272 } \\
    & MAE   & \textcolor{deepblue}{\em 0.453 } & \textcolor{deepblue}{\em 0.513 } & \textcolor{deepblue}{\em 0.571 } & \textcolor{deepblue}{\em 0.517 } & \textcolor{deepblue}{\em 0.576 } & \textcolor{deepblue}{\em 0.776 } & \textcolor{deepblue}{\em 0.294 } & \textcolor{deepblue}{\em 0.405 } & 0.836  & 0.504  & 0.521  & 0.549  & \textcolor{deepblue}{\em 0.304 } & \textcolor{deepblue}{\em 0.332 } & \textcolor{deepblue}{\em 0.361 } \\
    \hline
    \multicolumn{1}{|c|}{LightTS} & MSE   & \textbf{0.429 } & \textbf{0.466 } & \textbf{0.542 } & \textbf{0.416 } & \textbf{0.497 } & \textbf{0.739 } & \textbf{0.175 } & \textbf{0.272 } & \textbf{0.391 } & \textcolor{deepblue}{\em 0.511 } & \textcolor{deepblue}{\em 0.527 } & \textcolor{deepblue}{\em 0.554 } & \textbf{0.176 } & \textbf{0.219 } & \textbf{0.235 } \\
          & MAE   & \textbf{0.443 } & \textbf{0.468 } & \textbf{0.536 } & \textbf{0.448 } & \textbf{0.499 } & \textbf{0.610 } & \textbf{0.267 } & \textbf{0.335 } & \textbf{0.420 } & \textcolor{deepblue}{\em 0.495 } & \textcolor{deepblue}{\em 0.509 } & \textcolor{deepblue}{\em 0.525 } & \textbf{0.279 } & \textbf{0.318 } & \textbf{0.329 } \\
    \hline
    \multicolumn{1}{|c|}{Improvement}  &   MSE   & 4.67\% & 11.74\% & 9.21\% & 24.91\% & 24.35\% & 33.90\% & 11.17\% & 22.29\% & 34.18\% & -42.34\% & -7.11\% & -5.12\% & 11.11\% & 6.41\% & 13.60\% \\
    & MAE   & 2.21\% & 8.77\% & 6.13\% & 13.35\% & 13.37\% & 21.39\% & 9.18\% & 17.28\% & 24.60\% & -19.85\% & -3.67\% & -4.37\% & 8.22\% & 4.22\% & 8.86\% \\

    \hline
    \end{tabular}

    \begin{tablenotes}
    \item{Results are taken from \cite{Informer} (for results from LogTrans to Informer). Best result in one setting is marked in \textbf{bold}, and second best is marked in \textcolor{deepblue}{\em italic}. *We use 5 seeds to calculate the average result of SCINet and AutoFormer. We follow the same look-back settings in Informer and SCINet \cite{Informer, SCINet} for each dataset.}
    \end{tablenotes}
    
    \end{threeparttable}
  \label{tab:ETT}%
\end{table*}%

\begin{table*}[!t]\scriptsize
\renewcommand\arraystretch{1.6}
  \centering
  \caption{Baseline comparisons under single-step setting for short sequence time series forecasting tasks.}
  
  \begin{threeparttable}
    \begin{tabular}{|c|c|cccc|cccc|cccc|cccc|}
    \hline
    \multicolumn{1}{|c|}{Methods} & \multicolumn{1}{c|}{Metrics} & \multicolumn{4}{c|}{Solar-Energy} & \multicolumn{4}{c|}{Traffic}  & \multicolumn{4}{c|}{Electricity} & \multicolumn{4}{c|}{Exchange-Rate} \\
\cline{3-18}          &       & \multicolumn{4}{c|}{horizon}  & \multicolumn{4}{c|}{horizon}  & \multicolumn{4}{c|}{horizon}  & \multicolumn{4}{c|}{horizon} \\
\cline{3-18}          &       & 3     & 6     & 12    & 24    & 3     & 6     & 12    & 24    & 3     & 6     & 12    & 24    & 3     & 6     & 12    & 24 \\
    \hline
    \multicolumn{1}{|c|}{AR} & RSE   & 0.2435 & 0.3790 & 0.5911 & 0.8699 & 0.5991 & 0.6218 & 0.6252 & 0.6300  & 0.0995 & 0.1035 & 0.1050 & 0.1054 & 0.0228 & 0.0279 & 0.0353 & \textcolor{deepblue}{\em 0.0445} \\
          & CORR  & 0.9710 & 0.9263 & 0.8107 & 0.5314 & 0.7752 & 0.7568 & 0.7544 & 0.7591 & 0.8845 & 0.8632 & 0.8691 & 0.8595 & 0.9734 & 0.9656 & 0.9526 & 0.9357 \\
    \hline
    \multicolumn{1}{|c|}{VARMLP \cite{VARMLP}} & RSE   & 0.1922 & 0.2679 & 0.4244 & 0.6841 & 0.5582 & 0.6579 & 0.6023 & 0.6146 & 0.1393 & 0.1620 & 0.1557 & 0.1274 & 0.0265 & 0.0394 & 0.0407 & 0.0578 \\
          & CORR  & 0.9829 & 0.9655 & 0.9058 & 0.7149 & 0.8245 & 0.7695 & 0.7929 & 0.7891 & 0.8708 & 0.8389 & 0.8192 & 0.8679 & 0.8609 & 0.8725 & 0.8280 & 0.7675 \\
    \hline
    \multicolumn{1}{|c|}{GP \cite{GPTS}} & RSE   & 0.2259 & 0.3286 & 0.5200  & 0.7973 & 0.6082 & 0.6772 & 0.6406 & 0.5995  & 0.1500  & 0.1907 & 0.1621 & 0.1273 & 0.0239 & 0.0272 & 0.0394 & 0.0580 \\
          & CORR  & 0.9751 & 0.9448 & 0.8518 & 0.5971 & 0.7831 & 0.7406 & 0.7671 & 0.7909 & 0.8670 & 0.8334 & 0.8394 & 0.8818 & 0.8713 & 0.8193 & 0.8484 & 0.8278 \\
    \hline
    \multicolumn{1}{|c|}{RNN-GRU} & RSE   & 0.1932 & 0.2628 & 0.4163 & 0.4852 & 0.5358 & 0.5522 & 0.5562 & 0.5633 & 0.1102 & 0.1144 & 0.1183 & 0.1295 & 0.0192 & 0.0264 & 0.0408 & 0.0626 \\
          & CORR  & 0.9823 & 0.9675 & 0.9150 & 0.8823 & 0.8511 & 0.8405 & 0.8345 & 0.8300 & 0.8597 & 0.8623 & 0.8472 & 0.8651 & 0.9786 & \textbf{0.9712} & 0.9531 & 0.9223 \\
    \hline
    \multicolumn{1}{|c|}{LSTNet \cite{LSTNet}} & RSE   & 0.1843 & 0.2559 & 0.3254 & 0.4643 & 0.4777 & 0.4893 & 0.4950 & 0.4973 & 0.0864 & 0.0931 & 0.1007 & 0.1007 & 0.0226 & 0.0280 & 0.0356 & 0.0449 \\
          & CORR  & 0.9843 & 0.9690 & 0.9467 & 0.8870 & 0.8721 & 0.8690 & 0.8614 & 0.8588 & 0.9283 & 0.9135 & 0.9077 & 0.9119 & 0.9735 & 0.9658 & 0.9511 & 0.9354 \\
    \hline
    \multicolumn{1}{|c|}{TCN} & RSE   & 0.1940 & 0.2581 & 0.3512 & 0.4732 & 0.5459 & 0.6061 & 0.6367 & 0.6586 & 0.0892 & 0.0974 & 0.1053 & 0.1091 & 0.0217 & 0.0263 & 0.0393 & 0.0492 \\
          & CORR  & 0.9835 & 0.9602 & 0.9321 & 0.8812 & 0.8486 & 0.8205 & 0.8048 & 0.7921 & 0.9232 & 0.9121 & 0.9017 & 0.9101 & 0.9693 & 0.9633 & 0.9531 & 0.9223 \\
    \hline
    \multicolumn{1}{|c|}{TPR-LSTM} & RSE   & 0.1803 & 0.2347 & 0.3234 & 0.4389 & 0.4487 & 0.4658 & 0.4641 & 0.4765 & 0.0823 & 0.0916 & 0.0964 & 0.1006 & \textbf{0.0174} & \textbf{0.0241} & \textcolor{deepblue}{\em 0.0341} & \textbf{0.0444} \\
          & CORR  & 0.9850 & \textcolor{deepblue}{\em 0.9742} & 0.9487 & \textbf{0.9081} & 0.8812 & 0.8717 & 0.8717 & 0.8629 & 0.9439 & 0.9337 & 0.9250 & 0.9133 & \textcolor{deepblue}{\em 0.9790} & 0.9709 & \textbf{0.9564} & \textbf{0.9381} \\
    \hline
    \multicolumn{1}{|c|}{MTGNN} & RSE   & \textcolor{deepblue}{\em 0.1778} & 0.2348 & 0.3109 & 0.4270 & \textcolor{deepblue}{\em 0.4162} & 0.4754 & \textcolor{deepblue}{\em 0.4461} & 0.4535 & \textbf{0.0745} & 0.0878 & \textbf{0.0916} & \textbf{0.0953} & 0.0194 & 0.0259 & 0.0349 & 0.0456 \\
          & CORR  & \textcolor{deepblue}{\em 0.9852} & 0.9726 & 0.9509 & 0.9031 & \textbf{0.8963} & 0.8667 & \textbf{0.8794} & \textbf{0.8810} & \textcolor{deepblue}{\em 0.9474} & 0.9316 & 0.9278 & 0.9234 & 0.9786 & 0.9708 & \textcolor{deepblue}{\em 0.9551} & \textcolor{deepblue}{\em 0.9372} \\
    \hline
    \multicolumn{1}{|c|}{SCINet*} & RSE   & 0.1788 & \textcolor{deepblue}{\em 0.2319} & \textcolor{deepblue}{\em 0.3049} & \textcolor{deepblue}{\em 0.4249} & 0.4203 & \textcolor{deepblue}{\em 0.4447} & 0.4536 & \textcolor{deepblue}{\em 0.4477} & \textcolor{deepblue}{\em 0.0758} & \textbf{0.0852} & 0.0934 & \textcolor{deepblue}{\em 0.0973} & 0.0179 & 0.0249 & 0.0344 & 0.0462 \\
          & CORR  & 0.9849 & 0.9735 & \textcolor{deepblue}{\em 0.9529} & 0.9026 & \textcolor{deepblue}{\em 0.8931} & \textbf{0.8802} & \textcolor{deepblue}{\em 0.8760} & \textcolor{deepblue}{\em 0.8783} & \textbf{0.9493} & \textbf{0.9386} & \textcolor{deepblue}{\em 0.9296} & \textbf{0.9272} & 0.9744 & 0.9655 & 0.9493 & 0.9279 \\
    \hline
    \multicolumn{1}{|c|}{LightTS} & RSE   & \textbf{0.1704} & \textbf{0.2212} & \textbf{0.2930} & \textbf{0.4133} & \textbf{0.3973} & \textbf{0.4335} & \textbf{0.4403} & \textbf{0.4416} & 0.0762 & 0.0876 & 0.0935 & 0.0985 & \textcolor{deepblue}{\em 0.0178} & \textcolor{deepblue}{\em 0.0246} & \textbf{0.0339} & 0.0453 \\
          & CORR  & \textbf{0.9866} & \textbf{0.9761} & \textbf{0.9564} & \textcolor{deepblue}{\em 0.9065} & 0.8900 & \textcolor{deepblue}{\em 0.8731} & 0.8696 & 0.8699 & 0.9432 & 0.9304 & 0.9238 & 0.9191 & \textbf{0.9798} & \textcolor{deepblue}{\em 0.9710} & 0.9548 & 0.9360 \\
    \hline
    \multicolumn{1}{|c|}{Improvement} & RSE   & 4.16\% & 4.61\% & 3.90\% & 2.73\% & 4.54\% & 2.52\% & 1.30\% & 1.36\% & -2.28\% & -2.82\% & -2.07\% & -3.36\% & -2.30\% & -2.07\% & 0.59\% & -2.03\% \\
          & CORR  & 0.14\% & 0.20\% & 0.37\% & -0.18\% & -0.70\% & -0.81\% & -1.11\% & -1.26\% & -0.64\% & -0.87\% & -0.62\% & -0.87\% & 0.08\% & -0.02\% & -0.17\% & -0.22\% \\
    \hline
    \end{tabular}%
    \begin{tablenotes}
    \item {Results of previous models are taken from \cite{SCINet}. Best result in one setting is marked in \textbf{bold}, and second best is marked in \textcolor{deepblue}{\em italic}. *We use 5 seeds to calculate the average result of SCINet. We follow the same look-back settings as previous studies \cite{LSTNet,SCINet,MTGNN, TPR-LSTM} for each dataset.}
    
    \end{tablenotes}
    
    \end{threeparttable}
  \label{tab:financial}%
\end{table*}%

\subsection{Main Results}
\label{sec: main results}
Table \ref{tab:ETT} and Table \ref{tab:financial} show the main results of LightTS. We can observe that LightTS can achieve state-of-the-art or comparable results in most cases. In the following, we discuss the results of short sequence forecasting and long sequence forecasting, respectively.

\subsubsection{Long sequence forecasting}
Table \ref{tab:ETT} presents the experimental results on long sequence forecasting tasks. LightTS achieves state-of-the-art results over all the horizons in ETTh1, ETTh2, ETTm1, and Electricity datasets and achieves second best results in the Weather dataset. In particular, for the longest prediction horizon, LightTS lowers MSE by 9.21\%, 33.90\%, 34.18\%, and 13.60\% on the ETTh1, ETTh2, ETTm1, and Electricity datasets, respectively. Compared with transformer models (LogTrans, Reformer, Informer), RNN-based models (LSTNet, LSTMa), and CNN-based models (TCN, SCINet), LightTS achieves significant improvement in long sequence forecasting tasks. On the one hand, continuous and interval sampling can help the model capture the short-term local patterns and long-term global dependencies. On the other hand, such a down-sampling design is critical for the model to process a very long input sequence efficiently.

\subsubsection{Short sequence forecasting}
Table \ref{tab:financial} presents the experimental results on the task of short sequence forecasting. We can observe that LightTS achieves state-of-the-art results on the Solar-Energy dataset. In particular, on Solar-Energy dataset, LightTS lowers down RSE by 4.16\%, 4.61\%, 3.90\%, 2.73\% on horizon 3, 6, 12, 24 respectively. We also observe an inconsistency in the evaluation metrics of RSE and CORR. For example, on Traffic datasets, LightTS achieves state-of-the-art results in terms of RSE but slightly lags behind on CORR. For Traffic, Electricity, and Exchange-Rate datasets, none of the existing models can consistently outperform other models. LightTS provides comparable results to MTGNN and SCINet on these datasets. Moreover, as we will see in the next section, LightTS is efficient and has significant advantages over MTGNN and SCINet in FLOPS and running time.

\subsection{Comparisons on FLOPS and Running Time}
\label{sec:compute cost}
We compare the FLOPS and running time of LightTS with previous SOTA models (Autoformer, MTGNN, SCINet). We present the results of the three largest datasets in short sequence forecasting and four datasets in long sequence forecasting in Table \ref{tab:FLOPS} and \ref{tab:runningtime}. The hyperparameters of LightTS, MTGNN, and SCINet align with the ones that generate the forecasting results in Table \ref{tab:ETT} and \ref{tab:financial}. We can observe that LightTS has significant advantages in FLOPS and running time in both short sequence and long sequence forecasting. For the Traffic dataset (the largest dataset with the greatest number of variables), the FLOPS of LightTS is 96.2\% smaller than MTGNN and 99.4\% smaller than SCINet. In addition, LightTS achieves a speedup of 44.5x over MTGNN and 13.8x over SCINet in terms of running time per epoch. For the Electricity dataset in long sequence forecasting tasks, the FLOPS of LightTS is 93.5\% smaller than SCINet and 97.2\% smaller than Autoformer. We perform the experiments on a server with two 12-core Intel(R) Xeon(R) CPU E5-2690 v4 @ 2.60GHz and one Tesla V100 PCIe 16GB.

\begin{table}[t]
\renewcommand\arraystretch{1.3}
  \centering
  \caption{The FLOPS of Autoformer, MTGNN, SCINet and LightTS \cite{ptflops}. The best model is marked in \textbf{bold}. We report the FLOPS of the models corresponding to the longest forecasting horizon of each dataset in Table \ref{tab:ETT} and \ref{tab:financial}. Solar: Solar-Energy. ECL: Electricity.}
  \scalebox{0.83}{
  \begin{threeparttable}
      \begin{tabular}{|c|cccc|ccc|}
    \hline
    FLOPS (M) & \multicolumn{4}{c|}{Long Sequence} & \multicolumn{3}{c|}{Short Sequence} \\
\cline{2-8}          & ETTh1 & ETTm1 & Weather & ECL & Traffic & Solar & ECL \\
    \hline
    Autoformer & 7581  & 7581  & 7600  & 11652 & - & - & - \\
    MTGNN & - & - & - & - & 2370   & 377   & 883 \\
    SCINet & 6     & 17    & 15    & 5078  & 16348 & 205   & 57 \\
    \hline
    LightTS & \textbf{4} & \textbf{3} & \textbf{7} & \textbf{328} & \textbf{90} & \textbf{10} & \textbf{30} \\
    \hline
    \end{tabular}%
    \begin{tablenotes}
        \item - dash denotes that the method does not implement on this task.
    \end{tablenotes}
    \end{threeparttable}
  }
  \label{tab:FLOPS}%
\end{table}%

\begin{table}[t]
\renewcommand\arraystretch{1.3}
  \centering
  \caption{We report the running time of one epoch in seconds for Autoformer, MTGNN, SCINet and LightTS. We report the running time of each dataset with the longest forecasting horizon in Table \ref{tab:ETT} and \ref{tab:financial}. The experimental environment and batch size are the same for each model. The best model is marked in \textbf{bold}. Solar: Solar-Energy. ECL: Electricity.}
  \scalebox{0.83}{
  \begin{threeparttable}
      \begin{tabular}{|c|cccc|ccc|}
    \hline
    \multicolumn{1}{|c|}{Time (s)} & \multicolumn{4}{c|}{Long Sequence} & \multicolumn{3}{c|}{Short Sequence} \\
\cline{2-8}    \multicolumn{1}{|c|}{} & ETTh1 & ETTm1 & Weather & ECL & Traffic & Solar & ECL \\
    \hline
    \multicolumn{1}{|c|}{Autoformer} & 80    & 355   & 280   & 330   & -     & -     & - \\
    \multicolumn{1}{|c|}{MTGNN} & -     & -     & -     & -     & 1470  & 213   & 850 \\
    \multicolumn{1}{|c|}{SCINet} & 80    & 850   & 260   & 950   & 455   & 275   & 3750 \\
    \hline
    LightTS & \textbf{2} & \textbf{9} & \textbf{7} & \textbf{160} & \textbf{33} & \textbf{33} & \textbf{50} \\
    \hline
    \end{tabular}%
    \begin{tablenotes}
        \item - dash denotes that the method does not implement on this task.
    \end{tablenotes}
    \end{threeparttable}
  }
  \label{tab:runningtime}%
\end{table}%

\subsection{Robustness Analysis}
\begin{figure}[t]
    \centering
    \includegraphics[width=8cm]{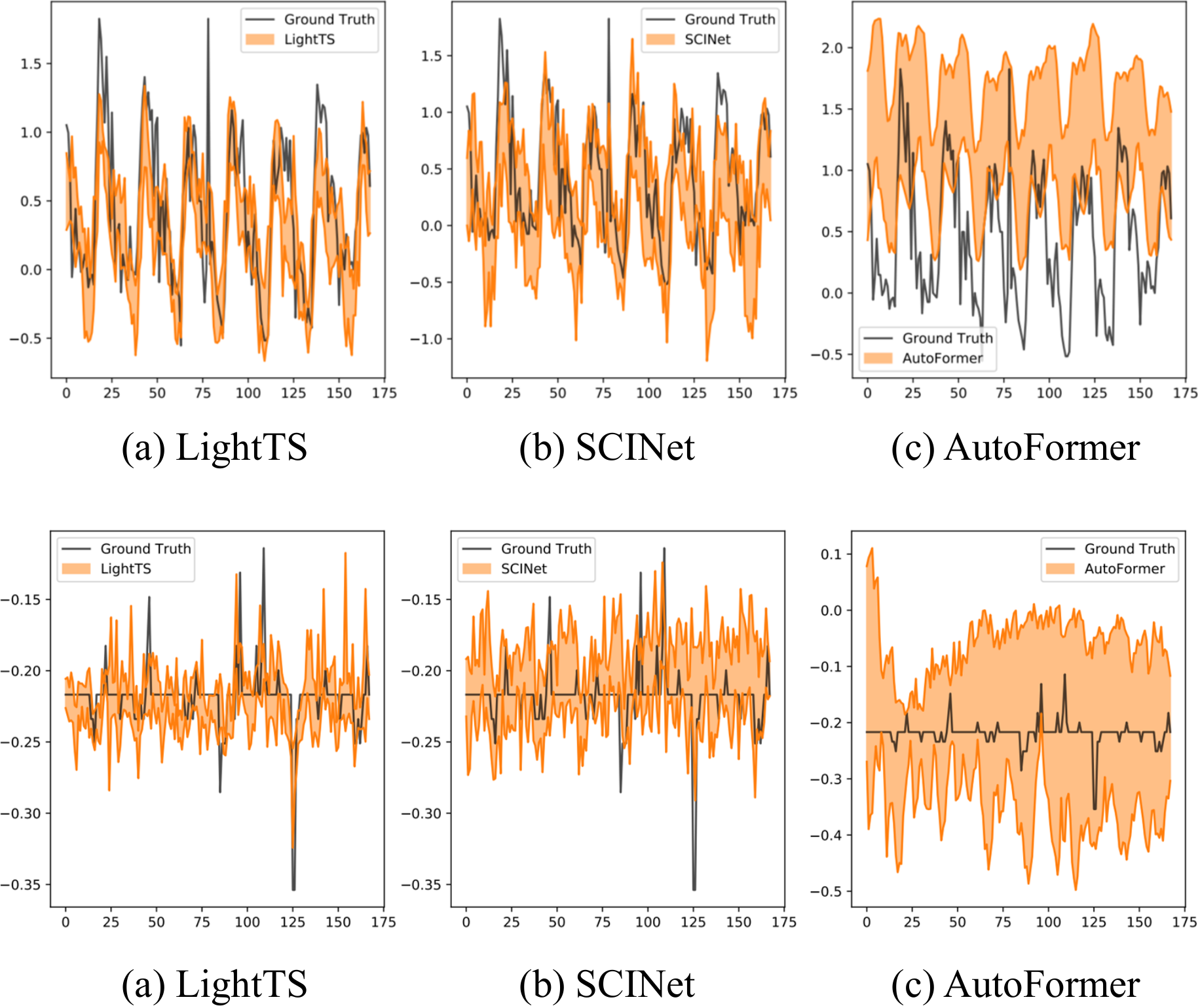}
    \caption{Forecasting sequence randomly selected from ETTh1 (upper row) and Electricity (lower row). The shaded area is the range of forecasting results under five different random seeds for training.}
    \label{fig: robustness}
\end{figure}

Robustness is a critical problem in long sequence time series forecasting. A wrong forecast on the trend or seasonality can gradually accumulate over time, resulting in incredibly misleading forecasting. During the experiments, we find that previous models for long sequence forecasting are not robust across random seeds, while LightTS can provide stable results. We present the standard deviation of AutoFormer, SCINet, and LightTS on the ETTh1, ETTm1, Weather, and Electricity datasets in Table \ref{tab: robustness}. We can observe that LightTS has a much smaller variance in prediction accuracy than AutoFormer and SCINet. We also present the shaded area of the forecasting by different random seeds for LightTS, SCINet, and AutoFormer in Figure \ref{fig: robustness}. We can observe that LightTS has a much smaller shaded area of forecasting across different random seeds than SCINet and AutoFormer. LightTS has a significant advantage over existing methods in long sequence forecasting from both accuracy and robustness.

\begin{table}[t]\small
\renewcommand\arraystretch{1.3}
  \centering
  
  \caption{The mean and standard deviation of the forecasting accuracy of LightTS, AutoFormer, and SCINet in long sequence forecasting. We report the results of the longest forecasting horizon for different datasets in Table \ref{tab:ETT}. The results are in the form of $mean_{std}$ estimated by five random seeds. LightTS has a much smaller variance in forecasting accuracy than AutoFormer and SCINet.}
  \scalebox{0.92}{
    \begin{tabular}{|c|c|c|c|c|c|}
    \hline
    \multicolumn{1}{|c}{Methods} & \multicolumn{1}{|c}{Metrics} & \multicolumn{1}{|c}{ETTh1}  & \multicolumn{1}{|c}{ETTm1} & \multicolumn{1}{|c}{Weather} & \multicolumn{1}{|c|}{Electricity} \\
    \hline
    \multicolumn{1}{|c|}{AutoFormer} 
        & MSE   & 0.898$_{0.039}$  & 0.599$_{0.075}$  & \textbf{0.527}$_{0.059}$ & 0.291$_{0.028}$  \\
        & MAE   & 0.743$_{0.019}$  & 0.542$_{0.027}$  & \textbf{0.503}$_{0.033}$ & 0.381$_{0.019}$  \\
    \hline
    \multicolumn{1}{|c|}{SCINet} 
        & MSE   & 0.597$_{0.013}$  & 1.214$_{0.274}$  & 0.577$\mathbf{_{0.003}}$  & 0.272$_{0.012}$  \\
        & MAE   & 0.571$_{0.010}$  & 0.836$_{0.104}$  & 0.549$\mathbf{_{0.002}}$  & 0.361$_{0.009}$ \\
    \hline
    \multicolumn{1}{|c|}{LightTS} 
        & MSE   & \textbf{0.498}$\mathbf{_{0.002}}$  & \textbf{0.358}$\mathbf{_{0.001}}$ & 0.554$\mathbf{_{0.003}}$  & \textbf{0.235}$\mathbf{_{0.003}}$ \\
        & MAE   & \textbf{0.512}$\mathbf{_{0.002}}$ &  \textbf{0.388}$\mathbf{_{0.001}}$ & 0.525$\mathbf{_{0.002}}$  & \textbf{0.329}$\mathbf{_{0.003}}$ \\
    \hline
    \end{tabular}%
    }
  \label{tab: robustness}%
\end{table}%

\begin{figure*}
    \centering
    \includegraphics[width=16cm]{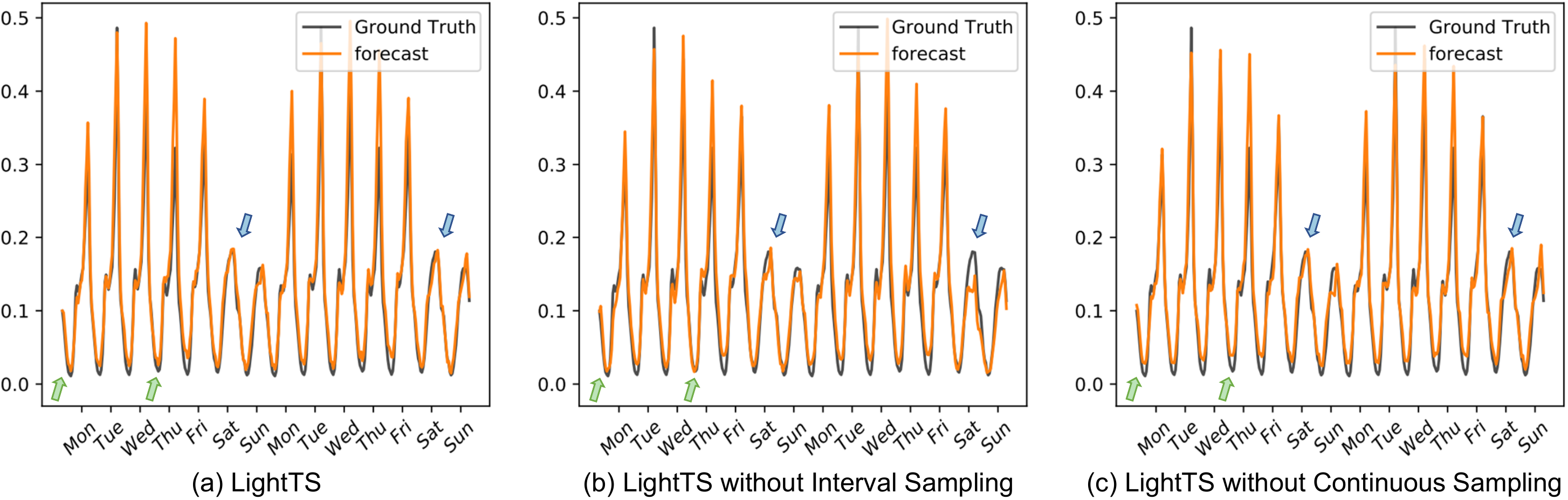}
    \caption{The ground truth (black) and the forecast (orange) for one variable in the Traffic dataset with the forecasting horizon of 24. The variable has daily patterns (short-term local patterns) and weekly patterns (long-range patterns). The green arrows point at daily patterns, where (a) and (b) which have continuous sampling are more accurate in forecasting than (c). The blue arrows point at weekly patterns, where (a) and (c) which have interval sampling are more accurate in forecasting than (b).}
    \label{fig: ablation}
\end{figure*}

\subsection{Ablation Study}
\label{sec:ablation study}
We conduct an ablation study to investigate the effectiveness of the components we propose in LightTS. We name LightTS without different components as follows:
\begin{itemize}
    \item \textbf{w/o CP}: LightTS without the channel projection. In this way, we do not rely on the interdependency of different time series to make predictions.
    \item \textbf{w/o IS}: LightTS without the interval sampling.
    \item \textbf{w/o CS}: LightTS without the continuous sampling.
\end{itemize}

We repeat each experiment 5 times with different random seeds. We report the mean and standard deviation of RSE and CORR over five runs in Table \ref{tab:ablation}. The introduction of channel projection significantly improves the performance of Solar and Electricity datasets. The interdependencies in these two datasets are evident. On the contrary, the interdependency in the Exchange-Rate dataset is not evident. The observation is similar in MTGNN \cite{MTGNN} where MTGNN also fails to achieve desirable results in Exchange-Rate. Removing channel projection improves the results in the Exchange-Rate dataset. When the interdependencies among different time series are not evident, forcing the model to capture interdependency may negatively impact the model's performance. The effects of interval and continuous sampling are evident for all the datasets. We also demonstrate how interval sampling and continuous sampling affect the forecasting of LightTS in Figure \ref{fig: ablation}. The results demonstrate that continuous sampling helps LightTS capture the short-term local patterns, and interval sampling helps LightTS capture the long-range patterns.

\begin{table}[h]
\renewcommand\arraystretch{1.3}
  \centering
  \small
  \caption{The results of the ablation study on different datasets. The forecasting horizon is set to 24 for each dataset. Best result in each setting is marked in bold. Solar: Solar-Energy. Exchange: Exchange-Rate.}
  \begin{threeparttable}
    \begin{tabular}{|c|c|c|c|c|c|}
    \hline
          \multicolumn{1}{|c|}{Methods} & \multicolumn{1}{c|}{Metrics} & \multicolumn{1}{c|}{Solar} & \multicolumn{1}{c|}{Traffic} & \multicolumn{1}{c|}{Electricity} & \multicolumn{1}{c|}{Exchange} \\
          
    \hline
    \multicolumn{1}{|c|}{LightTS} & RSE   & \textbf{0.4133} & \textbf{0.4416} & \textbf{0.0985} & 0.0453 \\
          & CORR  & \textbf{0.9065} & \textbf{0.8699} & 0.9191 & 0.9360 \\
    \hline
    \multicolumn{1}{|c|}{w/o CP} & RSE   & 0.4469 & 0.4489 & 0.0998 & \textbf{0.0441} \\
          & CORR  & 0.8912 & 0.8639 & 0.9119 & \textbf{0.9364} \\
    \hline
    \multicolumn{1}{|c|}{w/o IS} & RSE   & 0.4166 & 0.4571 & 0.1008 & 0.0486 \\
          & CORR  & 0.9062 & 0.8591 & 0.9126 & 0.9345 \\
    \hline
    \multicolumn{1}{|c|}{w/o CS} & RSE   & 0.4174 & 0.4425 & 0.0997 & 0.0475 \\
          & CORR  & 0.9044 & 0.8698 & \textbf{0.9193} & 0.9333 \\
    \hline
    \end{tabular}
    
    \end{threeparttable}
  \label{tab:ablation}%
\end{table}%

\section{Discussion: Study of the Channel Projection}
\label{sec: discussion}

One of the major challenges in multivariate time series forecasting is capturing the interdependency among different time series. In the design of LightTS, the channel projection in IEBlock C is the only module that communicates the information of different time series. In the implementation of LightTS, the channel projection is a simple linear layer. The question is, is a simple linear layer sufficient to model the interdependency among different time series? There are two difficulties in answering this question. The first difficulty is how to quantify the interdependency among different time series modeled by LightTS. The second difficulty is how to measure the quality of the interdependency modeled by LightTS.

\begin{figure}[h]
    \centering
    \includegraphics[width=0.45\textwidth]{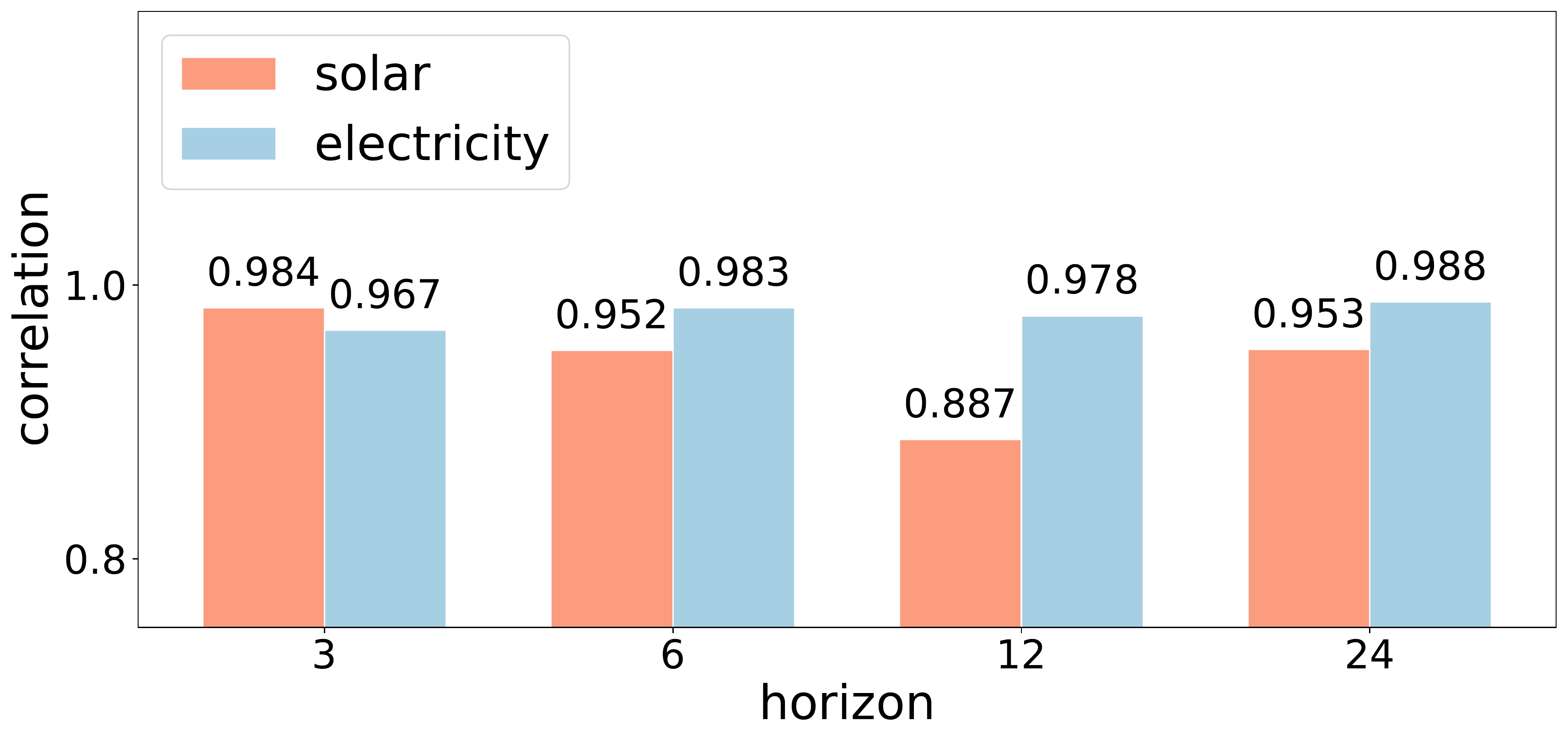}
    \caption{The correlation between the interdependency modeled by LightTS and MTGNN. The improvement of channel projection for Solar and Electricity is significant in our ablation study, which suggests that the interdependencies among different variables are evident in these two datasets.}
    \label{fig: correlation}
\end{figure}

For the first difficulty, we use the Deep SHAP (DeepLIFT + Shapley values) \cite{SHAP, DeepLIFT} to explain (and quantify) the interdependency among different time series in LightTS. Deep SHAP is a popular method that interprets the prediction of deep learning models based on DeepLIFT \cite{DeepLIFT} and SHAP \cite{SHAP}. For a given input matrix $X=(x_{ij})_{N\times T}$, we assume the predictions of LightTS are $\bm{\hat{y}}=(\hat{y_1},\hat{y_2},...,\hat{y_N})$. For the prediction of the $k$-th time series $\hat{y_k}$ ($k\in \{1,2,...,N\}$), Deep SHAP can explain how each element in the input matrix contributes to the prediction by LightTS. Assume the Deep SHAP values of the input $X$ are $S^k=(s_{ij}^k)_{N\times T}$, where $s_{ij}^k$ is the attribution of input $x_{ij}$ for prediction $\hat{y_k}$. Due to the fact that Deep SHAP is an additive feature attribution method \cite{SHAP}, the sum of the Deep SHAP values approximates the prediction \cite{SHAP}:
$
\hat{y_k} \approx \phi_0 + \sum_{i=1}^N\sum_{j=1}^T s_{ij}
$
where $\phi_0$ represents the model output with all the inputs missing \cite{SHAP}. Therefore, $s_i^k=\sum_{j=1}^Ts_{ij}^k$ evaluates how the input of time series $i$ contributes to the prediction $\hat{y_k}$ of time series $k$, in other words, how the prediction of time series $k$ by LightTS {\em depends} on time series $i$. We use the matrix $E=(\bm{e_1},\bm{e_2},...,\bm{e_N})$ to quantify the interdependency among different time series in LightTS, where $\bm{e_k}=(s_1^k,s_2^k,...,s_N^k)^T$. See Appendix \ref{app: SHAP Values} for more implementation details.

The question turns to how to measure the quality of the interdependency modeled by LightTS. The ``real'' interdependency among time series is not defined for many multivariate time series data. In this case, we use the interdependency modeled by MTGNN \cite{MTGNN} as the ``ground truth'' to measure the quality of the interdependency modeled by LightTS. MTGNN \cite{MTGNN} is a GNN-based method that explicitly learns the graph structure during training and is one of the state-of-the-art methods in multivariate time series forecasting. Following the above steps, we can also calculate matrix $M=(\bm{m_1}, \bm{m_2},...,\bm{m_N})$ to quantify the interdependency in MTGNN. We calculate the following metric:
$
correlation = \frac{1}{N}\sum_{i=1}^Ncorr(\bm{e_i}, \bm{m_i})
$
where $corr(\cdot, \cdot)$ is the Pearson correlation between two vectors. We present the results of 8 cases in our experiments in Figure \ref{fig: correlation}. We can see that the interdependency modeled by LightTS highly correlates with the one modeled by MTGNN. These empirical findings show that a simple channel projection is sufficient in learning the interdependency among different time series.

\section{Conclusion}
\label{sec: conclusion}
This paper proposes LightTS, a simple model merely based on multi-layer perceptrons. The key idea of LightTS is to apply an MLP-based architecture on top of two down-sampling strategies. Continuous and interval sampling helps the model capture the short-term local patterns and long-term temporal dependencies. Furthermore, since the model only needs to deal with a fraction of the input sequence after down-sampling, our model is highly efficient in handling time series with very long input sequences. Extensive experiments show that LightTS is accurate, efficient, and robust in short sequence and long sequence multivariate time series forecasting tasks.
\bibliographystyle{ACM-Reference-Format}
\bibliography{sample-base}

\newpage

\appendix

\section{Implementation Details}
\subsection{Datasets Description}
\label{app: datasets}

In Table \ref{tab:dataset}, we show basic information of datasets which we use to evaluate performance. Here we introduce details of these datasets.
\begin{itemize}
    \item \textbf{Electricity Transformer Temperature} (ETT): Introduced by \cite{Informer}, ETT datasets record 2-year electric power in two different counties in China. ETTm1 has a sample rate of 15 minutes, while ETTh1 and ETTh2 have a sample rate of 1 hour. These three datasets have seven variables in each timestamp. We use the three ETT datasets mentioned above to evaluate long sequence forecasting abilities. The train/val/test is 6/2/2.
    \item \textbf{Weather}: This dataset contains 4-year climatological data of multiple U.S. locations, in which data points are collected every 1 hour. This dataset has 12 variables in every timestamp. We use this dataset to evaluate long sequence forecasting abilities. The train/val/test is 7/1/2.
    \item \textbf{Electricity}: Electricity dataset (also known as Electricity Consuming Load, or ECL) collects electricity consumption of 321 users in two years (2012-2014) every 15 minutes. We use this dataset to evaluate long and short sequence forecasting abilities. In long sequence forecasting, the train/val/test is 7/1/2. In short sequence forecasting, the train/val/test is 6/2/2.
    \item \textbf{Solar-Energy}: This dataset (also called Solar) records the solar power production of 137 photovoltaic plants in Alabama State in 2006 every 10 minutes. We use this dataset to evaluate short sequence forecasting abilities. The train/val/test is 6/2/2.
    \item \textbf{Traffic}: This dataset collects two years (2015-2016) of hourly road occupancy rates data in California. Each timestamp contains 862 variables. We use this dataset to evaluate short sequence forecasting abilities. The train/val/test is 6/2/2.
    \item \textbf{Exchange-Rate}: This dataset collects daily exchange rates of eight countries, including Australia, Canada, China, Japan, New Zealand, Singapore, Switzerland, and the United Kingdom, from 1990 to 2016. We use this dataset to evaluate short sequence forecasting abilities. The train/val/test is 6/2/2.
    
\end{itemize}

\subsection{Evaluation Metrics}
\label{app:evaluation}

\subsubsection{Long Sequence Forecasting}

Following \cite{Informer}, we use Mean Squared Error (MSE) and Mean Absolute Error (MAE) defined as follows:
\begin{itemize}
    \item Mean Squared Error (MSE):
    $$ MSE = \frac{1}{n}\sum_{i=1}^n(Y-\hat{Y})^2$$
    \item Mean Absolute Error (MAE):
    $$ MAE = \frac{1}{n}\sum_{i=1}^n\left|Y-\hat{Y}\right|$$
\end{itemize}
where $Y, \hat{Y} \in \mathbb{R}^{l \times N}$ ($l$ is the output length, $N$ is the number of variables) are the ground truth and prediction, $n$ is the size of the test set. For both MSE and MAE, a lower value means better performance.

\subsubsection{Short Sequence Forecasting}

Following \cite{LSTNet}, we use Root Relative Squared Error (RSE), and Empirical Correlation Coefficient (CORR) defined as follows:
\begin{itemize}
    \item Root Relative Squared Error (RSE):
    $$RSE = \frac{\sqrt{\sum_{(t,i)\in\Omega_{Test}}{(Y_{ti}-\hat{Y}_{ti})}^2}}{\sqrt{\sum_{(t,i)\in\Omega_{Test}}{(Y_{ti}-mean(Y))}^2}}$$
    \item Empirical Correlation Coefficient (CORR):
    $$CORR = \frac{1}{n}\sum_{i=1}^n\frac{\sum_t(Y_{ti}-mean(Y_i))(\hat{Y}_{ti}-mean(\hat{Y}_i))}{\sqrt{\sum_t{(Y_{ti}-mean(Y_i))}^2{(\hat{Y}_{ti}-mean(\hat{Y}_i))}^2}}$$
\end{itemize}
where $Y, \hat{Y} \in \mathbb{R}^{l \times N}$ ($l$ is the output length, $N$ is the number of variables) are the ground truth and prediction, and $\Omega_{Test}$ is test set. For RSE, a lower value means better performance. For CORR, higher CORR means better performance.

\subsection{SHAP Values}
\label{app: SHAP Values}
In Section \ref{sec: discussion}, we use the Deep SHAP (DeepLIFT + Shapley values) \cite{SHAP, DeepLIFT} to explain the prediction of LightTS. We use 100 samples (input matrix) randomly selected from the training set as the background dataset for integrating out features. The SHAP value of each input element is its average SHAP value on 100 randomly selected test samples.

\section{Additional Experiment Results}

In this section, we show additional experiment results on long sequence forecasting in Table \ref{tab:additionresults}.

\begin{table*}[h]
  \centering
  \caption{Additional baseline comparisons under multi-step setting for long sequence time series forecasting tasks.}
  \begin{threeparttable}
    \begin{tabular}{|c|c|cc|cc|cc|cc|cc|}
    \hline
    \multicolumn{1}{|c|}{Methods} & \multicolumn{1}{c|}{Metrics} & \multicolumn{2}{c|}{ETTh1} & \multicolumn{2}{c|}{ETTh2} & \multicolumn{2}{c|}{ETTm1} & \multicolumn{2}{c|}{Weather} & \multicolumn{2}{c|}{Electricity} \\
\cline{3-12}          &       & \multicolumn{2}{c|}{horizon} & \multicolumn{2}{c|}{horizon} & \multicolumn{2}{c|}{horizon} & \multicolumn{2}{c|}{horizon} & \multicolumn{2}{c|}{horizon} \\
\cline{3-12}          &       & 24    & 48    & 24    & 48    & 24    & 48    & 24    & 48    & 48    & 168 \\
    \hline
    \multicolumn{1}{|c|}{LogTrans} & MSE   & 0.656  & 0.670  & 0.726  & 1.728  & 0.341  & 0.495  & 0.365  & 0.496  & 0.267  & 0.290  \\
          & MAE   & 0.600  & 0.611  & 0.638  & 0.944  & 0.495  & 0.527  & 0.405  & 0.485  & 0.366  & 0.382  \\
    \hline
    \multicolumn{1}{|c|}{Reformer} & MSE   & 0.887  & 1.159  & 1.381  & 1.715  & 0.598  & 0.952  & 0.583  & 0.633  & 1.312  & 1.453  \\
          & MAE   & 0.630  & 0.750  & 1.475  & 1.585  & 0.489  & 0.645  & 0.497  & 0.556  & 0.911  & 0.975  \\
    \hline
    \multicolumn{1}{|c|}{LSTMa} & MSE   & 0.536  & 0.616  & 1.049  & 1.331  & 0.511  & 1.280  & 0.476  & 0.763  & 0.388  & 0.492  \\
          & MAE   & 0.528  & 0.577  & 0.689  & 0.805  & 0.517  & 0.819  & 0.464  & 0.589  & 0.444  & 0.498  \\
    \hline
    \multicolumn{1}{|c|}{LSTNet} & MSE   & 1.175  & 1.344  & 2.632  & 3.487  & 1.856  & 1.909  & 0.575  & 0.622  & 0.279  & 0.318  \\
          & MAE   & 0.793  & 0.864  & 1.337  & 1.577  & 1.058  & 1.085  & 0.507  & 0.553  & 0.337  & 0.368  \\
    \hline
    \multicolumn{1}{|c|}{Informer} & MSE   & 0.509  & 0.551  & 0.446  & 0.934  & 0.325  & 0.472  & 0.353  & 0.464  & 0.269  & 0.300  \\
          & MAE   & 0.523  & 0.563  & 0.523  & 0.733  & 0.440  & 0.537  & 0.381  & 0.455  & 0.351  & 0.376  \\
    \hline
    \multicolumn{1}{|c|}{Autoformer*} & MSE   & 0.408  & 0.443  & 0.302  & 0.364  & 0.150  & 0.216  & \textbf{0.175 } & \textbf{0.224 } & 0.183  & 0.210  \\
          & MAE   & 0.434  & 0.451  & 0.374  & 0.417  & 0.264  & 0.315  & \textbf{0.259 } & \textbf{0.305 } & 0.299  & 0.325  \\
    \hline
    \multicolumn{1}{|c|}{SCINet*} & MSE   & \textcolor{deepblue}{\em 0.353 } & \textcolor{deepblue}{\em 0.389 } & \textcolor{deepblue}{\em 0.188 } & \textcolor{deepblue}{\em 0.339 } & \textcolor{deepblue}{\em 0.128 } & \textcolor{deepblue}{\em 0.157 } & \textcolor{deepblue}{\em 0.322 } & 0.421  & \textcolor{deepblue}{\em 0.151 } & \textcolor{deepblue}{\em 0.171 } \\
          & MAE   & \textcolor{deepblue}{\em 0.385 } & \textcolor{deepblue}{\em 0.411 } & \textcolor{deepblue}{\em 0.287 } & \textcolor{deepblue}{\em 0.400 } & \textcolor{deepblue}{\em 0.231 } & \textcolor{deepblue}{\em 0.265 } & \textcolor{deepblue}{\em 0.346 } & 0.431  & \textcolor{deepblue}{\em 0.252 } & \textcolor{deepblue}{\em 0.275 } \\
    \hline
    \multicolumn{1}{|c|}{LightTS} & MSE   & \textbf{0.314 } & \textbf{0.355 } & \textbf{0.178 } & \textbf{0.251 } & \textbf{0.105 } & \textbf{0.139 } & 0.326  & \textcolor{deepblue}{\em 0.387 } & \textbf{0.140 } & \textbf{0.150 } \\
          & MAE   & \textbf{0.356 } & \textbf{0.384 } & \textbf{0.269 } & \textbf{0.326 } & \textbf{0.197 } & \textbf{0.235 } & 0.351  & \textcolor{deepblue}{\em 0.402 } & \textbf{0.244 } & \textbf{0.254 } \\
    \hline
    \multicolumn{1}{|c|}{Improvement} & MSE   & 11.05\% & 8.74\% & 5.32\% & 25.96\% & 17.97\% & 11.46\% & -86.29\% & -72.77\% & 7.28\% & 12.28\% \\
          & MAE   & 7.53\% & 6.57\% & 6.27\% & 18.50\% & 14.72\% & 11.32\% & -35.52\% & -31.80\% & 3.17\% & 7.64\% \\
    \hline
    \end{tabular}%
     \begin{tablenotes}
    \item Results are taken from \cite{Informer} (for results from LogTrans to Informer). Best result in one setting is marked in \textbf{bold}, and second best is marked in \textcolor{deepblue}{\em italic}. *We use 5 seeds to calculate the average result of SCINet and AutoFormer. We follow the same look-back settings in Informer and SCINet \cite{Informer, SCINet} for each dataset.
    
    \end{tablenotes}
    
    \end{threeparttable}
  \label{tab:additionresults}%
\end{table*}%

\end{document}